\documentclass{bmvc2k}

\usepackage{multirow}
\usepackage{booktabs}
\def\eg{\textit{e.g.}~}
\def\ie{\textit{i.e.}~}

\def\etal{\textit{et al.}~}

\usepackage{amssymb}
\usepackage{flushend}
\usepackage{epstopdf}
\usepackage{subfigure}

\title{Geometry-Aware Video Object Detection for Static Cameras}

\addauthor{Dan Xu}{danxu@robots.ox.ac.uk}{1}
\addauthor{Weidi Xie}{weidi@robots.ox.ac.uk}{1}
\addauthor{Andrew Zisserman}{az@robots.ox.ac.uk}{1}

\addinstitution{
 Visual Geometry Group,\\
 Department of Engineering Science,\\ 
 University of Oxford,\\
 Oxford, UK
}

\runninghead{Xu, Xie, Zisserman}{Geometry-Aware Video Object Detection}

\def\eg{\emph{e.g}\bmvaOneDot}

\def\etal{\emph{et al}\bmvaOneDot}

\begin{document}
\maketitle

\begin{abstract}

In this paper we propose a geometry-aware model for video object detection. Specifically, we consider the setting that cameras can be well approximated as static, \eg in video surveillance scenarios, and scene pseudo depth maps can therefore be inferred easily from the object scale on the image plane.

We make the following contributions: 
\emph{First},
we extend the recent anchor-free detector~(CornerNet~\cite{Law18}) to video object detections.
In order to exploit the spatial-temporal information while maintaining high efficiency, 
the proposed model accepts video clips as input, and only makes predictions for the starting and the ending frames, \ie heatmaps of object bounding box corners and the corresponding embeddings for grouping.
\emph{Second}, 
to tackle the challenge from scale variations in object detection, scene geometry information, \eg derived depth maps, 
is explicitly incorporated into deep networks for multi-scale feature selection and for the network prediction.
\emph{Third},
we validate the proposed architectures on an autonomous driving dataset generated from the 
Carla simulator~\cite{Dosovitskiy17},
and on a real dataset for human detection~(DukeMTMC dataset~\cite{ristani2016performance}). 
When comparing with the existing competitive single-stage or two-stage detectors, 
the proposed geometry-aware spatio-temporal network achieves significantly better results. 
\end{abstract}


\section{Introduction}
\label{sec:intro}
\begin{figure*}[t]
\centering
\subfigure[A False Positive Detection Case]{\includegraphics[width=0.325\textwidth]{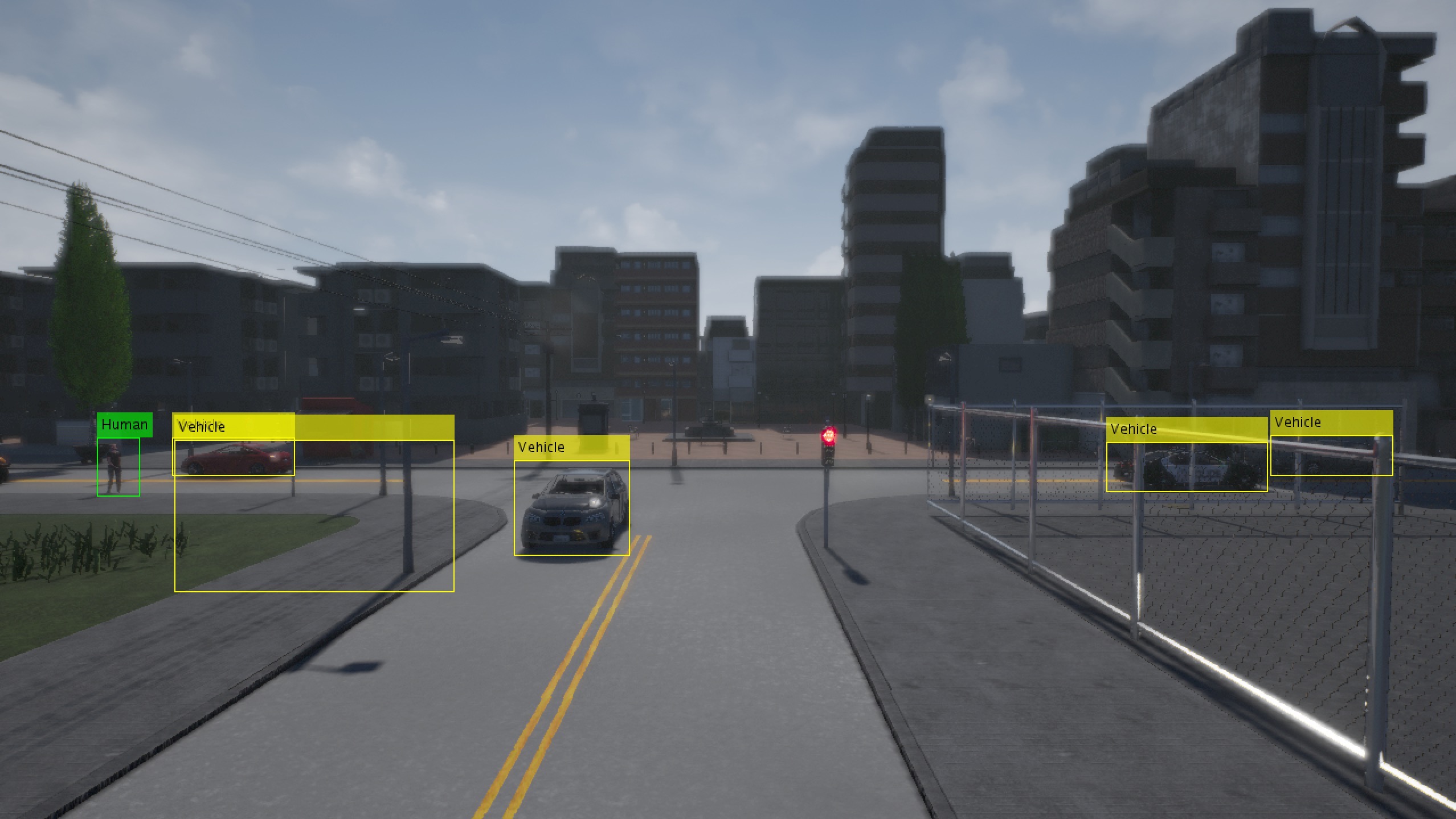}}
\subfigure[Height in Pixels of Objects]{\includegraphics [width=0.325\linewidth]{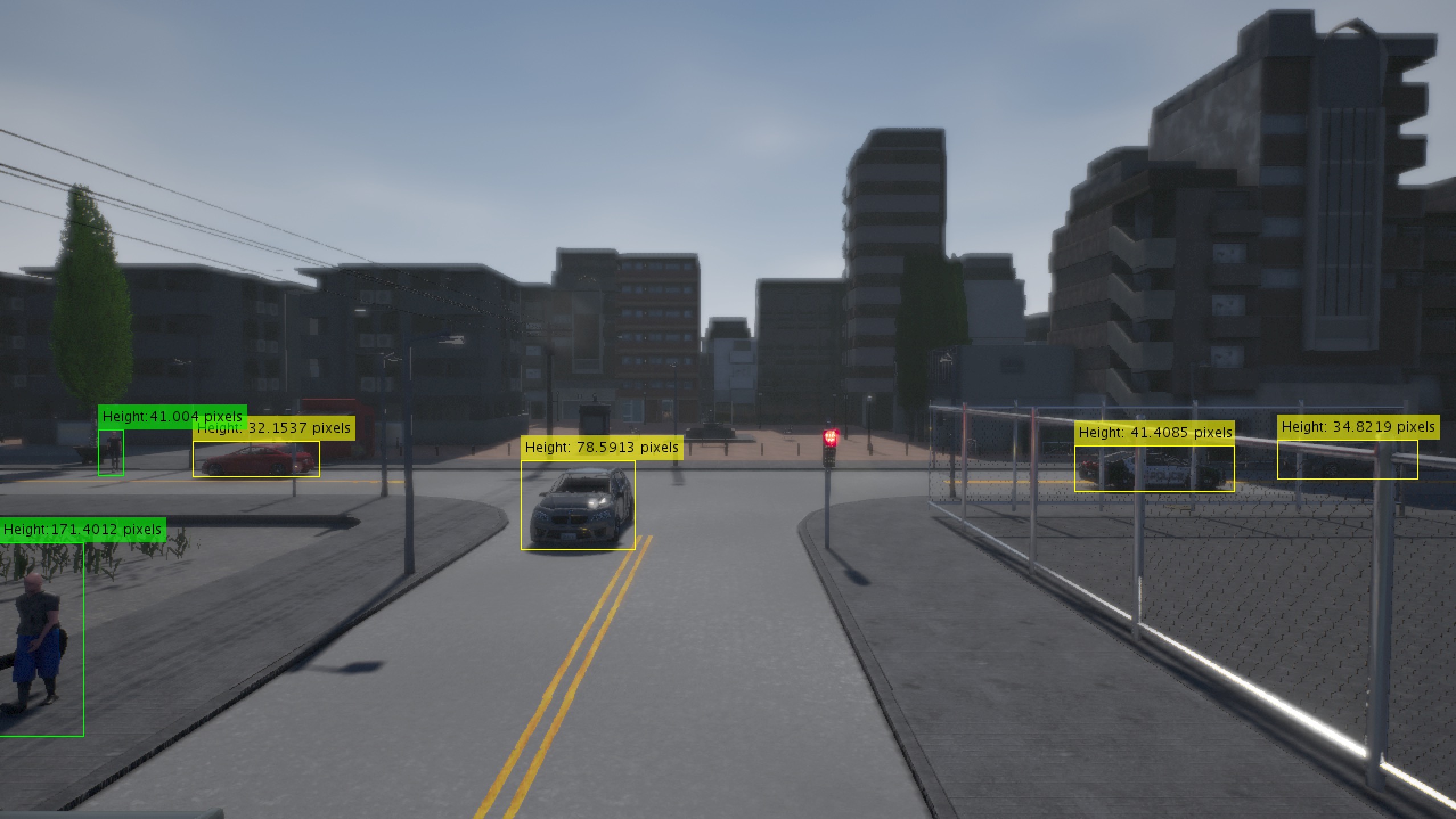}}
\subfigure[Pseudo Depth Map of Humans]{\includegraphics [width=0.325\linewidth]{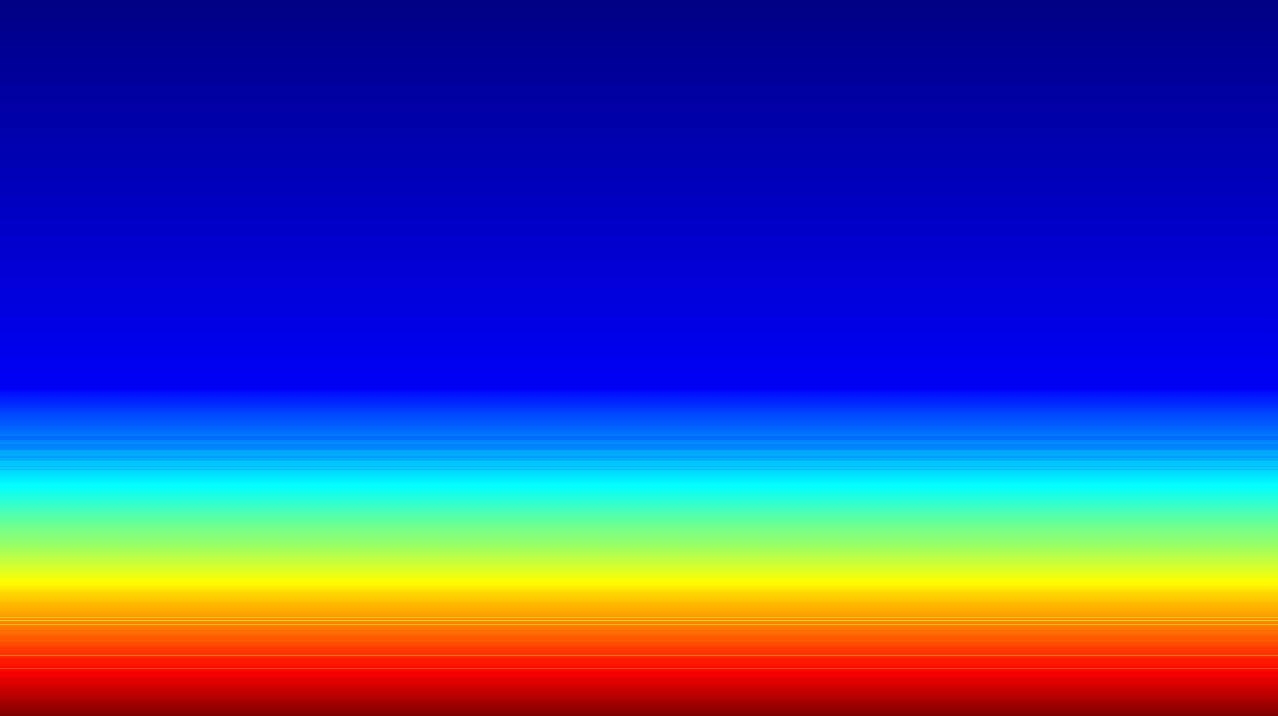}}
\caption{An illustration of the motivation of using scene geometry for detection: (a) a false positive detection of vehicle with a wrong scale; (b) the height of objects in different classes; (c) the geometry priors derived from (b), \ie 2.5D pseudo depth maps, able to provide useful geometric constraints on the object scales for learning a geometry-aware detector.}
\label{motivation}
\vspace{-10pt}
\end{figure*}

In the Deep Learning era, 
we always expect the deep networks to learn all the required world knowledge given sufficient training data. 
However, 
as images are essentially a 2D projection of the 3D world, 
and the depth information has been lost during the image formation~\cite{hartley2003multiple},
the scene geometry, \eg the depth, 
plays an essential role in resolving the ambiguities from scale variations and object occlusion in images.
Despite the great success achieved in video object detection~\cite{han2016seq, tripathi2016context,kang2016object,lu2017online}, 
detecting objects under different scales or occlusions has only partially been tackled via learning multi-scale features in a brute-force fashion~\cite{chen2016attention} or utilizing aggressive data augmentation~\cite{pepikj2013occlusion}.
This may indicate that these detectors actually have not learnt the scene geometry well.

These existing approaches mostly work on videos collected from dynamic cameras or internet video streams, 
such as the ImageNet VID dataset~\cite{ILSVRC15}, and thus the varying scene geometry is difficult to 
incorporate explicitly.
However, there are many real-world applications, for instance the video surveillance, 
where the cameras can be well-approximated as static settings, \ie the camera sits at a fixed position,
and the relative depth of an object in the world coordinate system can also be determined by its height on the image plane~\cite{Hoiem06}.
In this paper, we propose a deep model for video object detection under static cameras,
where the relative scene depth information can be estimated effectively~(as shown in Figure~\ref{motivation}), 
and is further used as strong geometric constraint for learning scale-aware object detector.

\par To investigate the effectiveness of using scene geometry in video object detection under the static camera settings, 
we first design a compact video object detector. 
While the two-stage anchor-based object detectors~(\eg Faster-RCNN~\cite{Ren16}) 
have achieved impressive accuracy on image-wise object detection, 
the efficiency of the model is usually upper-bounded by the region proposal process. 
The generation of proposals would also significantly increase the complexity of the model 
especially when we deal with object detection in videos. 
%
In this paper, 
we extend the more efficient single-stage anchor free and single-frame object detection model CornerNet~\cite{Law18} by incorporating spatio-temporal information, 
the proposed geometry-aware spatio-temporal network is termed as GAST-Net.
Our main contribution is therefore three-fold: 
(i) we design a spatio-temporal corner network structure, which accepts video clips~(image sequences) as input. 
As far as we know, this is the first use of a corner-based scheme for object detection in videos. 
The network utilizes a spatio-temporal convolutional backbone to encode appearance and motion representations, 
which are further used to predict and group corners only for the first frame and the last frame of the sequences. 
By doing so, we are able to capture the long temporal dependencies in videos.
(ii) We explore how the scene geometry derived from static cameras can be employed as priors for multi-scale feature selection and for the network prediction, 
therefore helping to tackle the challenges from scale variations in video object detection. 
(iii) Extensive experiments have been conducted on a synthetic automous driving dataset generated with Carla~\cite{Dosovitskiy17}, 
and on pedestrian detection on the DukeMTMC dataset~\cite{ristani2016performance}.
On both datasets, we demonstrate great benefits of incorporating the scene geometry, 
and show that the proposed GAST-Net significantly outperforms existing competitive single-stage and two-stage detectors. 

\section{Related Work}
\par\noindent\textbf{Object Detection in Static Images.}
Two families of detectors are currently popular: 
First, two-stage detectors, 
\eg R-CNN~\cite{Girshick13}, Fast R-CNN~\cite{Girshick15}, Faster R-CNN~\cite{Ren16} and R-FCN~\cite{Dai16}. 
The main idea of these detectors is to train a small sub-network for generating proposals that potentially contain objects, 
and then learn a classification network to predict the existence and categories of the objects.
Second, one-stage detectors that predict object bounding boxes in one step such as YOLO~\cite{Redmon16}, SSD~\cite{Liu16} and CornerNet~\cite{Law18}. 
Our model is also an one-stage detector which first predicts corner heatmaps and embeddings, 
then groups the corners as individual objects similar to CornerNet, 
while we extend it by building up a spatio-temporal corner network to capture temporal information for video object detection. 

\par\noindent\textbf{Object Detection in Videos.}
As an important research topic, video object detection has drawn significant attention~\cite{vu2018tube, tripathi2016context, prest2012learning, kwak2015unsupervised, joulin2014efficient, tang2019object, xiao2018video, bertasius2018object}. 
To take advantage of existing image-based detectors, 
several works focus on post-processing class scores from image-based detectors, 
and enforce temporal consistency on the scores. For instance, tubelet proposals are generated in~\cite{kang2016object} via applying a tracker to frame-based bounding box proposals. The class scores along the tubelet are further re-scored by a 1D CNN model.
Unlike the detection problem in static images, videos naturally contain temporal coherence with objects changing smoothly in time. 
Zhu~\etal~\cite{Zhu17} thus consider a motion-based model that applies a detection net only on key frames,
and an optical flow net is used for propagating deep features to the rest of the frames. 
To further simplify dense prediction in optical flow, 
a recent work~\cite{Feichtenhofer17} proposes to simultaneously learn detection and tracking.
In this paper, we consider a common case of video object detection, 
where the camera is static.
Under this situation, all the derived geometry information can therefore be applied to help the CNN with geometry-aware learning, 
in order to eliminate the scale ambiguities in 2D images. 

\par\noindent\textbf{Object Detection from RGB-D data.} 
Another line of research is about using RGB-D data where the scene depth information has been demonstrated beneficial for various computer vision tasks~\cite{schwarz2018rgb,lin2013holistic}, 
and is also widely used for object detection~\cite{gupta2014learning,eitel2015multimodal,spinello2011people}. 
Among the existing works, 
Gupta~\etal~\cite{gupta2014learning}~proposed a joint framework for object detection and semantic segmentation, 
and the depth maps are encoded with a geocentric encoding approach to provide complementary features to the RGB representations.  Spinello~\etal~\cite{spinello2011people} also explored using RGB-D data as input for people detection. There also exists some works exploring using depth data for 3D object detection~\cite{song2016deep,qi2018frustum}. Qi~\etal~\cite{qi2018frustum} developed a 3D point cloud deep representation model for effective 3D object bounding box prediction. 
However, these works require the explicit depth data captured from depth sensors, 
which are not always available in many application scenarios. 
Our work targets deriving the scene depth information from the RGB data, and thus does not require additional depth sensors other than standard RGB cameras.

\par\noindent\textbf{Geometry-Aware Deep Learning.} 
Scene geometry is considered as important prior information for computer vision tasks~\cite{wang2018geometry, Liu15}. 
Leibe~\etal~\cite{leibe2008coupled} explored joint object tracking and detection using geometry assumptions within a traditional non-deep-learning framework. 
In crowd counting~\cite{Liu15}, as the camera usually sits on a fixed position and the variance between people's height is small, 
it is easy to obtain the homography between the image and the head plane. 
By incorporating this information in the model, it becomes possible to directly predict the crowd density in the physical world.
Moreover, previous works~\cite{Hoiem06} also considered to place the local object detection in the context of the overall 3D scene,
by directly modelling the interdependence of objects, surface orientations, and camera viewpoint. 
In our work, we aim to explore using scene geometry for the task of video object detection in CNN. 
Instead of estimating accurate 3D geometry, 
we consider deriving and utilizing scene-specific coarse depth as well as image-plane coordinates, 
and enforce the convolutional operations to be conditioned on the object scales and positions, 
leading to a geometry-aware deep learning. 

\section{Geometry-Aware Spatio-Temporal Corner Network} 
\begin{figure}[!t]
  \centering
  \includegraphics[width=\textwidth]{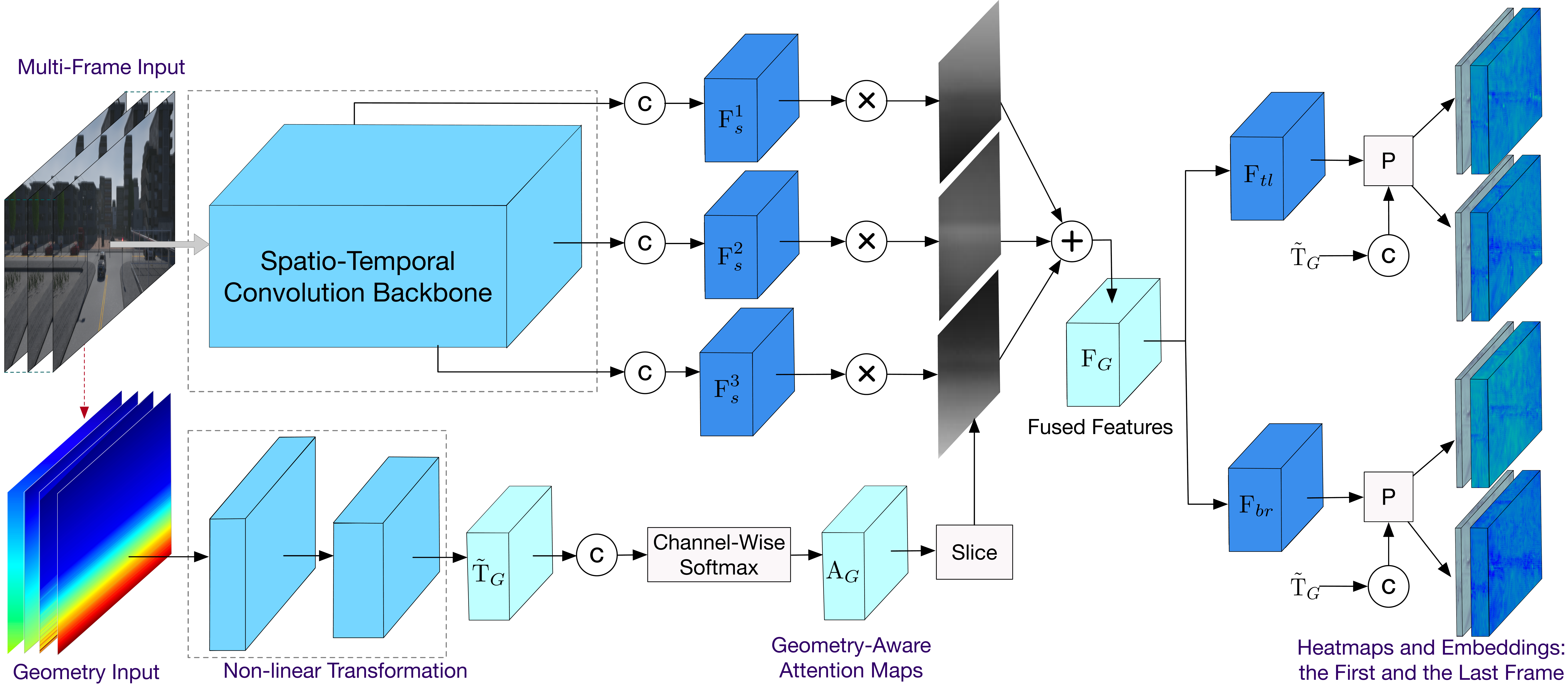}
  \caption{Framework of the proposed geometry-aware spatio-temporal corner network for video object detection from static cameras. It accepts multiple frames as input, and predicts heatmaps and embeddings of the first and the last frame for detection. The geometry input contains 2D image-plane coordinates and 2.5D pseudo depth maps, which can be directly derived from training data. $\textbf{P}$ denotes a prediction module. The symbols $\textcircled{c}$, $\otimes$ and $\oplus$ denote convolution, element-wise multiplication and element-wise addition operation, respectively. } %
\label{fig:framework}
\end{figure}

Figure~\ref{fig:framework} depicts a framework overview of the proposed GAST-Net. 
It consists of two main components. 
The first is the proposed spatio-temporal network that accepts video clips as input,
and outputs multi-heads feature representations with both appearance and motion information at different scales.
The second component is a geometry-aware module that first encodes the inferred relative depth maps (\ie the pseudo depth maps), 
and further used for selecting the features dynamically based on the geometry information. 
Intuitively, given the depth for all pixels on the image plane, 
in order to detect objects that are close to the camera, 
features from a large receptive field should be used. 
In our case, 2D image-plane coordinates and the pseudo depth maps are used to represent the image and scene geometry, which can both be derived from the training data.
Eventually, the corner heatmaps and embeddings are predicted from the fused feature representation, 
and bounding boxes are obtained by grouping the corners.
We introduce the details of the proposed GAST-Net in the following.
\subsection{Spatio-Temporal Corner Network}
\par\noindent\textbf{Single-Frame Corner Network.} 
In contrast to the traditional anchor-based detectors, such as SSD~\cite{Liu16} and Faster-RCNN~\cite{Ren16}, 
CornerNet~\cite{Law18} is an one-stage and anchor free detector in a bottom-up fashion. The main idea is to regress heatmaps for the top-left and the bottom-right corners of the objects and predict their embeddings. 
The bounding boxes that outline the objects are later generated via grouping corners from the embeddings. 
\vspace{5pt}
\par\noindent\textbf{Proposed Multi-Frame Spatio-Temporal Corner Network.} 
For efficient object detection in videos, we extend the CornerNet architecture by exploiting the spatio-temporal information.
To capture the temporal relationship between adjacent frames, 
the input to our model is video clips with multiple frames, and 
a 3D convolution-based backbone is used as an encoder~(variants of VGG and ResNet50 in our experiments). 
The video clip representation $\mathrm{F}_G$~(Figure.~\ref{fig:framework}) is further projected to two separate feature maps 
$\mathrm{F}_{tl}$ and $\mathrm{F}_{br}$~(corresponding to the first and last frame respectively), 
and later decoded as heatmaps and embeddings of the top-left and bottom-right corners for the object bounding boxes.
In order to improve the efficiency and reduce the computational overhead, supervision is only added on the first and the last frame. 
Multi-scale context is enabled by taking feature maps from different layers from the encoder, 
and the spatio-temporal representation $\mathrm{F}_G$ is calculated by fusing the multi-scale features with the proposed geometry-aware network module.

\subsection{Geometry-Aware Network Module}\label{geometry-aware}
\par\noindent\textbf{Scene Geometry from Static Cameras.} 
Static cameras are used in a wide range of real-world applications, where an important task is to detect and track all the cars and pedestrians.
In these scenarios, as the variations of the objects' physical height tend to be small, 
the depth information will be directly related to their sizes on image plane. 
For instance, the further objects will have smaller scales according to the perspective projection of the camera~\cite{Hoiem06}. 
Therefore, the geometry information from static cameras \ie the relative depth, can be directly estimated from the training data, as shown the psudo depth map in Figure~\ref{motivation}.

In this work, we mainly consider two types of geometry information, \eg image geometry and scene geometry. 
The image geometry considers the image-plane 2D coordinates as auxiliary information,
which can be treated as a means to enable position-dependent convolutions.
We generate a set of two coordinate maps \{$\mathrm{G}_x$, $\mathrm{G}_y$\} for the $x$ and the $y$ dimension respectively. 
In $\mathrm{G}_x \in \mathbb{R}^{H\times W}$, with $H$ and $W$ as the height and the width of the images, 
each column is the $x$ dimension coordinate, 
while in $\mathrm{G}_y \in \mathbb{R}^{H\times W}$, 
each row is the $y$ dimension coordinate. 
$\mathrm{G}_x$ and $\mathrm{G}_y$ are normalized in the range of [0,1]. 
The other type of geometry information is the scene geometry, \ie relative scene depth maps. 
For a fixed camera viewpoint $v_m$, 
the object height and its depth are inversely proportional~\cite{Hoiem06}. 
Given the bounding boxes of all the objects from the training data, 
we are able to estimate a coarse relative depth map by calculating the mean of the maximal and the minimal height of the bounding boxes for each row on the map. 
The rows without any objects are bilinearly interpolated using the values of adjacent rows. For each object class $c_n$ and each camera viewpoint $v_m$, 
we estimate such a pseudo depth map, and for the whole training data, 
we have a set of class- and scene-specific pseudo depth maps, \{$\mathrm{D}_{v_m, c_n} \in \mathbb{R}^{H\times W}\}(m=1,...,M, n=1,...,N)$, where $N$ and $M$ are the number of object classes and camera viewpoints, respectively.
\vspace{5pt}
\par\noindent\textbf{Non-Linear Transformation of Geometry Information.} 
Given a camera view $v_m$, we concatenate all the coordinate maps with the pseudo depth maps,
and perform non-linear transformations with two Convolution-Batch Norm-ReLU blocks denoted as $\mathrm{Conv {\text -} BN {\text -} ReLU}_2(\cdot)$. 
Then the transformation operation is formulated as follows:
\begin{align}
\tilde{\mathrm{T}}_G = \mathrm{Conv} {\text -} \mathrm{BN} {\text -} \mathrm{ReLU}_2(\mathrm{concat}(\mathrm{G}_x, \mathrm{G}_y, \mathrm{D}_{v_m, c_1}, ..., \mathrm{D}_{v_m, c_N})),
\end{align}
where $\mathrm{concat}(\cdot)$ is a concatenation operation. 
After that, we obtain a fine-grained geometry distribution $\tilde{\mathrm{T}}_G$. 
In our framework, $\tilde{\mathrm{T}}_G$ is used to guide the multi-scale feature selection with an attention mechanism, 
and is also used to guide the prediction of heatmaps and embeddings for later grouping corners. 

\vspace{5pt}
\par\noindent\textbf{Geometry-Aware Multi-Scale Feature Fusion.} 
In order to detect objects of different scales in the image plane,
the geometry information is used to modulate the multi-scale features with an attentional process.
Given a set of $S$ multi-scale features $\{F_s^i\}_{i=1}^S$, 
we correspondingly learn a set of $S$ geometry-aware attention maps $\mathrm{A}_G$, $A_G = \{A_G^i\}_{i=1}^S$.
Our intuition of using $\tilde{\mathrm{T}}_G$ for attention generation is that the geometry information, \eg pseudo depth map, essentially has strong constraints to the object scales on image plane. 
Formally, we generate the attention maps as follows:
\begin{align}
\mathrm{A}_G = \mathrm{Softmax}(W_G \tilde{\mathrm{T}}_G + b_G), 
\end{align}
$\mathrm{Softmax}(\cdot)$ is computed along the channels, 
and $\{W_G, b_G\}$ are the convolution parameters. 
Then the set of attention maps $\{A_G^i\}_{i=1}^S$ is used to select and fuse features in different scales as follows:
\begin{align}
\mathrm{F}_G = \mathrm{A}_G^1 \otimes \mathrm{F}_s^1 + \cdots + \mathrm{A}_G^S \otimes \mathrm{F}_s^S,
\end{align}
where the symbol $\otimes$ denotes an element-wise multiplication operation.

\vspace{5pt}
\par\noindent\textbf{Geometry-Aware Prediction.} 
The geometry distribution $\tilde{\mathrm{T}}_G$ is also used to guide the prediction of heatmaps and embeddings in the spatio-temporal corner network. 
The prediction part has four independent convolutional layers, 
corresponding to the top-left and bottom-right corners of the first frame and the last frame, respectively. 
For each prediction layer, it accepts features from both the image sub-network and from the geometry sub-network, 
\ie~$\tilde{\mathrm{T}}_G$. 
A separate $1\times 1$ convolution is applied on $\tilde{\mathrm{T}}_G$ to adjust the feature dimensions. 
In our setting, the number of geometry feature channels is set to $1/4$ of the image one. 
Then these two parts of features are concatenated and input into the prediction convolutional layer.

\subsection{Network Optimization and Inference} 
The overall network architecture uses a combination of two types of losses. One is a regression loss on the corner heatmaps, 
and the other is an embedding loss for both the first frame and the last frame. Similar to~\cite{Law18}, the heatmap regression uses a focal loss, 
since the number of corner pixels is much fewer  than background pixels. 
The embedding loss employs a pull-push loss that aims to train the network to group the corners by a pull loss, 
and to separate the corners by a push loss. 
During the inference, we supply the network with testing video clips and the geometry input derived in the training phase, 
where we assume that the training and the testing data are collected under the same camera scenes, 
which is usually a common setting in applications with static cameras.   
For each frame in the videos, 
it could be the first frame or the last frame of the input clips, and thus each frame is predicted twice. 
We collect grouped bounding boxes from the two-times predictions and apply an NMS operation to get the final bouding box output of that frame.

\section{Experiments}
\begin{figure}[!t]
  \centering
  \includegraphics[width=\textwidth]{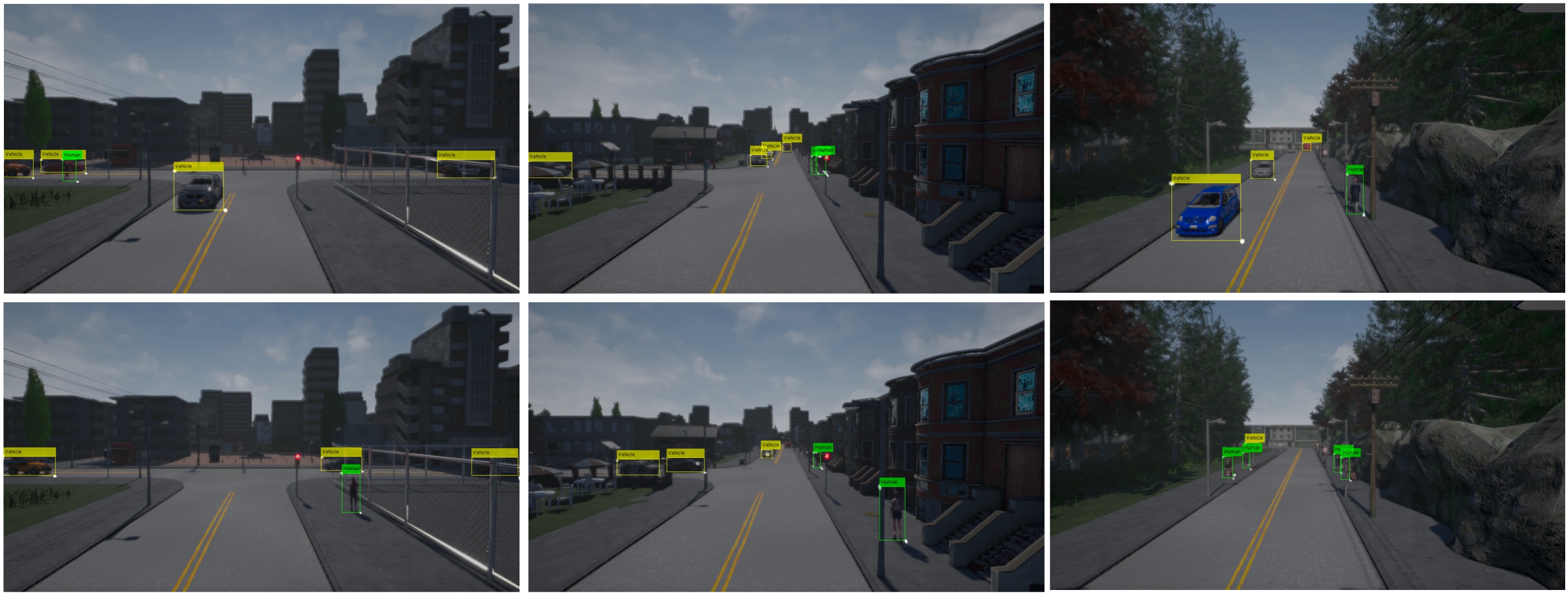}
  \caption{Qualitative detection results of humans and vehicles under three different scene views on Carla-Vehicle-Pedestrian dataset. The detected corners are visualized as grey blobs.} %
\label{fig:carla_qualitative}
\end{figure} 

\vspace{-5pt}
\subsection{Experimental Setup}\label{setup}
\begin{figure}[!t]
  \centering
  \includegraphics[width=\textwidth]{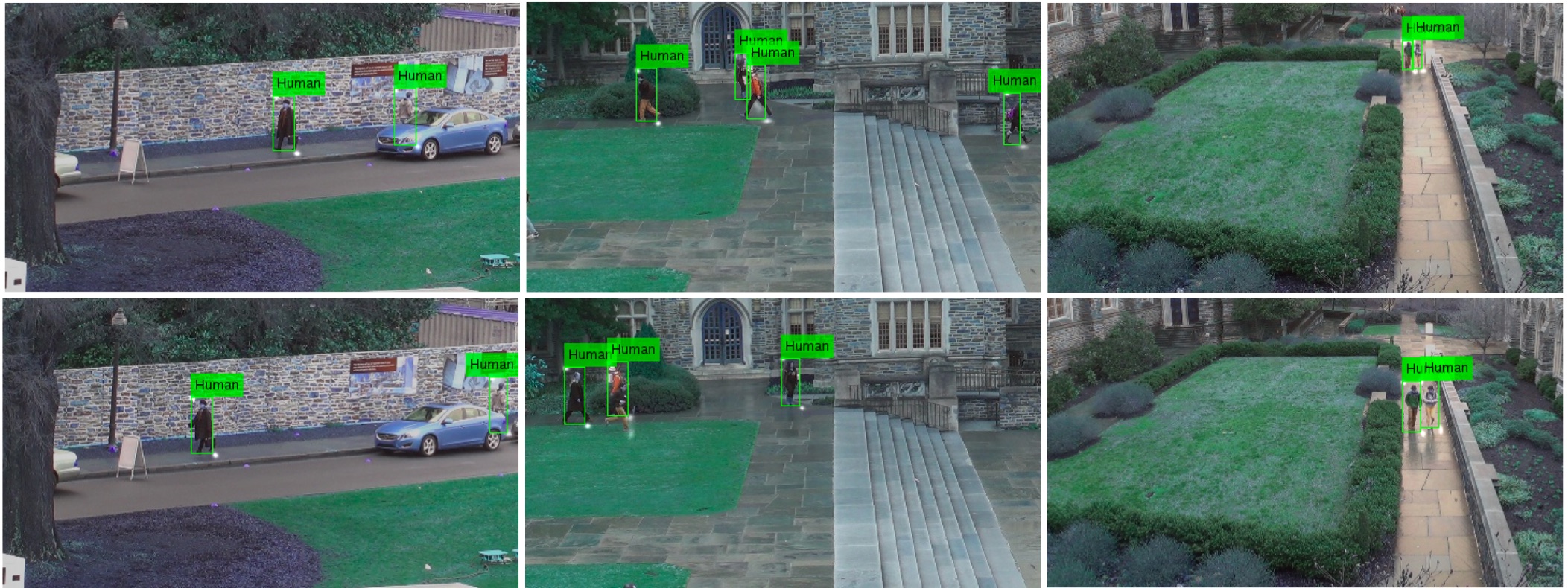}
  \caption{Qualitative detection results of humans under three different camera view points on the DukeMTMC dataset. 
  The detected corners are visualized in grey blobs.} %
\label{fig:duke_qualitative}
\end{figure}

\paragraph{Datasets.} 
We conduct the experiments on two different datasets: 
(i) a synthetic dataset generated from an open-sourced self-driving simulator Carla~\cite{Dosovitskiy17}, 
termed as Carla-Vehicle-Pedestrian dataset. 
We collected around 48 scenes with in total 60K images. 
The resolution of each image is of $720\times 1280$. 
Among them 40 scenes with around 50k images are used for training and the rest for testing. 
The dataset contains two classes of pedestrian and vehicle. 
This dataset is very challenging as it has been generated with many small-scale pedestrians and vehicle objects. 
The frame rate of this dataset is round 9 fps.
(ii) the DukeMTMC~\cite{ristani2016performance} dataset that was 
originally created for object tracking and person identification. 
The dataset contains long videos with 9 different static cameras. 
We use the video data from camera 1 to 5, 
and create a dataset of around 720K images. 
The frame rate of the DukeTMTC is 60 fps.
The image resolution is $1080\times 1920$. 
Among them 70\% are used for training and the rest for testing. 
In the training, we sample images at every 6th frame. 
Several qualitative detection samples on the two datasets are shown in Figure~\ref{fig:carla_qualitative} and Figure~\ref{fig:duke_qualitative}.

\vspace{5pt}
\par\noindent\textbf{Parameter Setting and Evaluation Metrics.} 
In training, the images are resized to a resolution of $360\times 640$ for Carla-Vehicle-Pedestrian, 
and $270\times 480$ for DukeMTMC. 
The number of input frames is set to 4 for both datasets. 
The batch size is set to 8 and 16, and the network is trained with 30 and 20 epochs for the two datasets respectively. 
We used Adam~\cite{KingmaB14} for optimization;
and the weights for the  regression focal loss, the push loss,  and the pull loss,  are set as 1, 0.1 and 0.1 respectively. 
The learning rate is initialized as $10^{-4}$ for both datasets. 
The detection performance is evaluated with the metric of average precision at IoU 0.5 (AP$^{50}$) and at IoU 0.75 (AP$^{75}$), 
and also with mAP, which is calculated by taking the average over the two APs. 

\subsection{Experimental Results}
\paragraph{Baseline models.} 
To demonstrate the effectiveness of different components in the proposed GAST-Net, 
we conduct experiments on several different models: 
(i) Single-Frame CornerNet, 
which we follow~\cite{Liu16} while replacing their hourglass backbone with a VGG-11 structure for fair comparison; 
(ii) GAST-Net (multi-frame), which is our base spatio-temporal corner network. 
We use a conv-3D network structure (\eg C3D~\cite{Tran15}) as the spatio-temporal convolutional backbone. 
The representations from the backbone are used to separately decode for the first and the last frame. 
This model does not employ any geometry information.
(iii) Single-Frame CornerNet w/ 2D coordinates or 2.5D pseudo depth map in prediction, 
which uses 2D coordinates or 2.5D pseudo depth map in the network prediction module via the geometry network branch, 
and using the encoded $\tilde{\mathrm{T}}_G$ to help the network prediction as we described in the Sec.~\ref{geometry-aware}. 
This baseline model is built upon the Single-Frame CornerNet, \ie the model (i); 
(iv) GAST-Net (multi-frame) w/ 2D coordinates or 2.5D pseudo depth map in prediction, 
which uses 2D coordinates or 2.5D pseudo depth map in the prediction module in a means similar to (iii) via utilizing the encoded geometry distribution $\tilde{\mathrm{T}}_G$ for the network prediction. This model is directly built upon the model GAST-Net (multi-frame), \ie the model (ii); 
(v) GAST-Net (multi-frame) w/ geometry-guided feature fusion~(our full model). 
It further uses the geometry information to guide the multi-scale feature fusion upon the model (iv) that uses the geometry only for the network prediction. All the models are learned in the same training setting as described in Sec.~\ref{setup} for a fair performance comparison.

\begin{figure*}[!t]
\centering
\subfigure[PR-Curve at IoU 0.5 on Vehicle Class]{\includegraphics[width=0.49\textwidth]{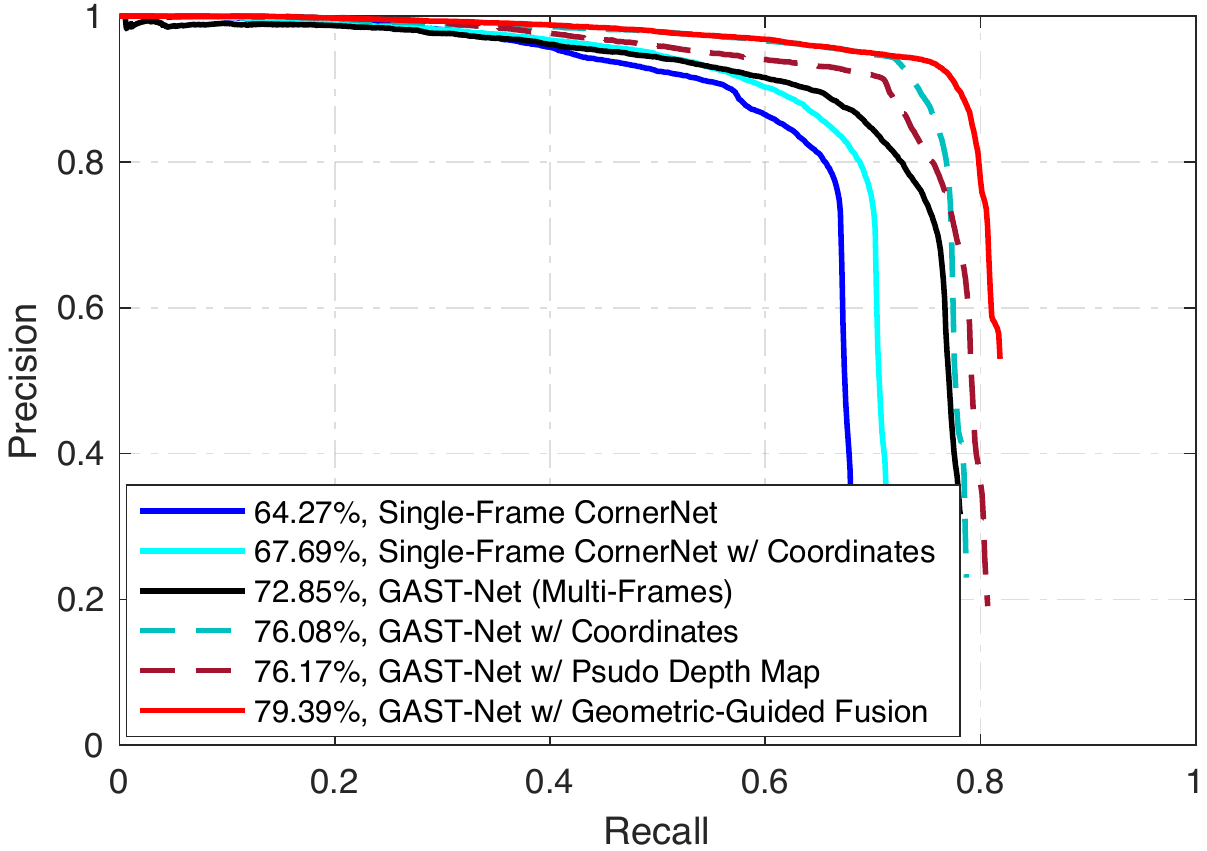}}
\subfigure[PR-Curve at IoU 0.5 on Pedestrian Class]{\includegraphics [width=0.49\linewidth]{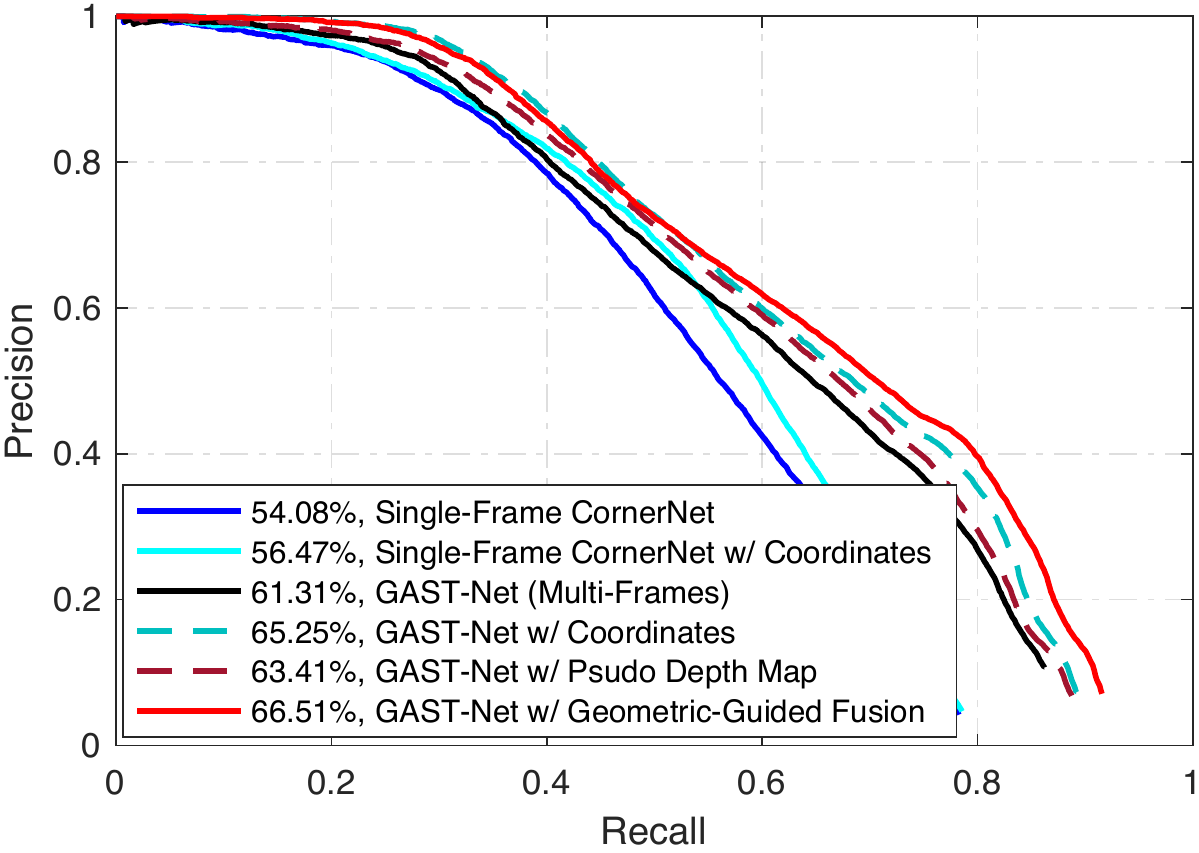}}
\caption{Comparison of Precision-Recall Curves of different variants of the proposed approach on the Carla-Vehicle-Pedestrian dataset.}
\label{PR_Curve}
\end{figure*}

\newcommand{\tabincell}[2]{\begin{tabular}{@{}#1@{}}#2\end{tabular}}
\begin{table}[!t]
\renewcommand{\arraystretch}{1.2} 
\addtolength{\tabcolsep}{-1pt}  
\centering
\footnotesize
\begin{tabular}{c|cccc}
\toprule
\multirow{2}{*}{Method} & \multicolumn{2}{c}{Vehicle-Class} & \multicolumn{2}{c}{Pedestrian-Class} \\ \cmidrule{2-5}
& AP$^{50}$ & AP$^{75}$ & AP$^{50}$ & AP$^{75}$ \\ \midrule\midrule
Single-Frame Corner Net~\cite{Liu16}  & 64.27\% & 52.20\% & 54.08\%& 28.63\% \\
GAST-Net (multi-frame) & 72.85\% & 62.37\% & 61.31\%& 54.82\% \\\midrule
Single-Frame Corner Net w/ 2D coordinates in prediction & 67.69\%& 55.83\% & 56.47\% &  31.56\%  \\
Single-Frame Corner Net w/ 2.5D psudo depth map  in prediction & 67.91\% & 54.75\% & 56.67\% &  30.83\% \\\midrule
GAST-Net (multi-frame) w/ 2D coordinates in prediction & 76.08\% & 69.02\% & 65.25\%&  57.54\% \\
GAST-Net (multi-frame) w/ 2.5D psudo depth map in prediction & 76.17\% & 66.52\% & 63.41\%& 56.11\% \\
GAST-Net (multi-frame) w/  geometry-guided feature fusion &  \textbf{79.39}\% & \textbf{71.95}\% & \textbf{66.51}\%& \textbf{59.06}\% \\
\bottomrule
\end{tabular}
\vspace{5pt}
\caption{Quantitative comparison of different variants of the proposed approach on the Carla-Vehicle-Pedestrian dataset. We use a backbone structure of C3D~\cite{Tran15}, which utilizes a VGG-11 structure while replacing  all 2D convolution/pooling with 3D convolution/pooling.}
\label{carla_ablation_study}
\vspace{-8pt}
\end{table}

\vspace{3pt}
\par\noindent\textbf{Effectiveness of multi-frame spatio-temporal corner prediction.} 
We conduct ablation study on the Carla-Vehicle-Pedestrian dataset. 
A quantitative comparison of different baseline models is shown in~Table~\ref{carla_ablation_study}. 
PR-Curves of the different approaches are shown in Figure~\ref{PR_Curve}. 
It can be observed that GAST-Net (multi-frame) significantly outperforms Single-Frame CornerNet on all the metrics by a large margin. 
In terms of AP$^{50}$, GAST-Net (multi-frame) is around 8.6 and 7.0 points better than Single-Frame CornerNet on the vehicle and the pedestrian class, respectively. 
The performance gain is even higher on the more strict metric of AP$^{75}$, 
demonstrating the effectiveness of incorporating temporal relationship in the proposed video object detection architecture.

\vspace{3pt}
\par\noindent\textbf{Effectiveness of geometry guided prediction.} 
When comparing the performance of GAST-Net (multi-frame) w/ 2D coordinates or 2.5D pseudo depth map with GAST-Net (multi-frame), 
it is clear that the geometry priors, \ie both the 2D image-plane coordinates and the 2.5D pseudo depth maps, 
are beneficial for improving the detection performance. 
On the vehicle class, GAST-Net with 2D coordinates improves AP$^{75}$ around 6.7 points, meaning that the coordinates are especially beneficial for the network to learn better localization of corners. 
We also use the geometry information for Single-Frame CornerNet, and consistent performance gains can be observed.

\vspace{3pt}
\par\noindent\textbf{Effectiveness of geometry guided multi-scale feature fusion.} 
In this section, we use the learned geometry distribution to guide multi-scale feature fusion.
As shown from Table~\ref{carla_ablation_study}, 
GAST-Net w/ Geometry-guided fusion further achieves better performance than model (iv) on all the metrics and on all the classes, verifying our initial motivation of encoding the geometry information into deep network for geometry-aware scale perception and learning.

\begin{table}[!t]
\renewcommand{\arraystretch}{1.2} 
\addtolength{\tabcolsep}{-1pt}  
\centering
\footnotesize
\begin{tabular}{c|c|ccc}
\toprule
Method & Backbone &  mAP & AP$^{50}$ & AP$^{75}$\\ \midrule\midrule
Faster RCNN~\cite{Ren16} & VGG & 63.85\% & {80.56}\% & 47.15\% \\\midrule
Single-Shot Detector (SSD)~\cite{Liu16} & VGG & 59.06\% & 73.87\% & 44.26\% \\
Single-Frame Corner Net~\cite{Law18} & VGG & 61.49\% & 72.65\%& 50.34\% \\
GAST-Net (full model) &  VGG & \textbf{68.26\%} & \textbf{78.65\%} & \textbf{57.87\%} \\\midrule\midrule
Faster RCNN~\cite{Ren16} & ResNet-50 & 70.68\% & {81.73}\%& 59.64\% \\\midrule
Single-Frame Corner Net~\cite{Law18}  & ResNet-50 & 68.71\% & 75.18\%& 62.25\% \\
GAST-Net (full model) &  ResNet-50 & \textbf{74.42\%} & \textbf{80.64\%} & \textbf{68.21\%} \\
\bottomrule
\end{tabular}
\vspace{5pt}
\caption{Quantitative comparison with competitive one-stage and two-stage detectors on the DukeMTMC dataset. Among the comparison methods, Faster RCNN~\cite{Ren16} is a two-stage anchor-based detector, while the rest are all one-stage detectors.}
\label{state-of-the-art}
\vspace{-8pt}
\end{table}

\vspace{5pt}
\par\noindent\textbf{Comparison with existing one-stage and two-stage detectors.}
We compare the proposed architecture with representative one-stage and two-stage object detectors, 
including Single-Shot MultiBox Detector (SSD)~\cite{Liu16}, Faster-RCNN~\cite{Ren16}, 
and Single-Frame CornerNet~\cite{Law18} on the DukeMTMC dataset. 
The comparison experiments are performed with two differnet backbone network structures, \eg VGG-11 and ResNet50. 
Quantitative comparisons are shown in Table~\ref{state-of-the-art}. 
GAST-Net achieves the best performance among these competitors. 
Specifically, to compare with the one-stage detectors, ours is 7.1 points better than Single-Frame CornerNet, 
and 8.1 points better than SSD on the mAP metric with VGG backbone. 
Ours is also around 4.4 points better than the two-stage Faster-RCNN approach. 
It can be also noted that, 
our corner-based framework has much better performance than anchor-based SSD and Faster-RCNN on AP$^{75}$, 
which is probably because that the dense prediction of corners is more powerful in accurate bounding box localization than using sparse anchor based proposal generation.

\begin{figure}[!t]
  \centering
  \includegraphics[width=\textwidth]{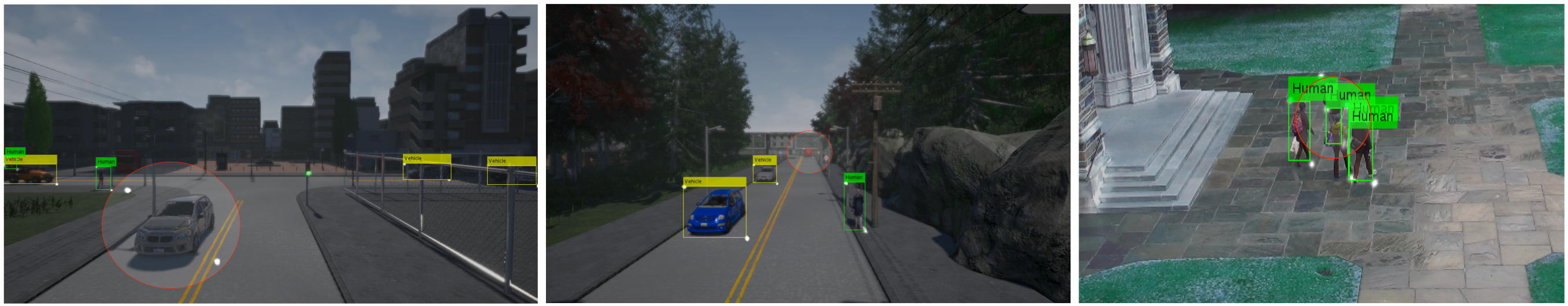}
  \caption{Failure examples on the CVP and the DukeTMTC datasets. The object detections that fail are marked with red circles. The missing (\eg the first two examples),  or inaccurate {\em grouping} (\eg the last example) of the detected top-left and bottom right corners is an important factor affecting the final detection performance.} %
\label{fig:failurecase}
\vspace{-5pt}
\end{figure} 

\vspace{3pt}
\par\noindent\textbf{Discussion.} 
The proposed GAST-Net is an one-stage based approach, 
which detects the top-left and the bottom-right corners, 
and learns to group the corresponding corners to bounding boxes. 
The final detection performance is thus affected by the grouping capability. 
In our experiments, we observed that the detector is able to produce very good detection and localization on the object corners \textit{w.r.t} the Percentage of Correct Keypoints (PCK) recall metric. 
However, the grouping fails in some cases, 
for instance, for  extreme scale of objects,  or for crowded cases  with dense occlusion, as shown in Figure~\ref{fig:failurecase}, 
leading to lower recall on the object bounding boxes. 
Possible solutions to tackle the grouping issues could be investigating a scale-aware network structure with long-term tracking. 

\vspace{-5pt}
\section{Conclusion}
We have presented a geometry-aware spatio-temporal network (GAST-Net)
for video object detection from static cameras.  GAST-Net consists of
two main parts.  One is the spatio-temporal corner network that aims
to perform object detection from corner estimation and grouping with
video clips as input.  The other part is the designed geometry-aware
network module which utilizes the scene geometry derived from static
cameras for multi-scale feature selection and fusion.  Extensive
experiments on two challenging datasets demonstrate the superior
performance of the proposed approach, and show the great advantage of
using geometry in deep networks for the video object detection
task. The  geometry-aware network module is also potentially
beneficial for other computer vision tasks affected by 
scale issues, such as object tracking and semantic segmentation.

\vspace{10pt}
\par\noindent\textbf{Acknowledgement.} This research work was supported by the EPSRC Programme Grant Seebibyte EP/M013774/1.

\bibliography{bmvc}
\end{document}